\documentclass[11pt]{article}
\usepackage{graphicx}
\usepackage{longtable,lscape}
\usepackage{amsfonts}
\usepackage{float}
\usepackage{url}

\newcommand{\captionfonts}{\footnotesize}
\makeatletter  
\long\def\@makecaption#1#2{%
  \vskip\abovecaptionskip
  \sbox\@tempboxa{{\captionfonts #1: #2}}%
  \ifdim \wd\@tempboxa >\hsize
    {\captionfonts #1: #2\par}
  \else
    \hbox to\hsize{\hfil\box\@tempboxa\hfil}%
  \fi
  \vskip\belowcaptionskip}
\makeatother 
\begin{document}
\title{Quantum Interaction Approach in Cognition, \\ Artificial Intelligence and Robotics}
\author{Diederik Aerts$^1$, Marek Czachor$^{1,2}$ and Sandro Sozzo$^1$
		\vspace{0.4 cm} \\ 
        \normalsize\itshape
        $^1$ Center Leo Apostel for Interdisciplinary Studies \\
        \normalsize\itshape
        Brussels Free University, Pleinlaan 2, 1050 Brussels, 
       Belgium \\
        \normalsize
        E-Mails: \url{diraerts@vub.ac.be,ssozzo@vub.ac.be}
        \vspace{0.2 cm} \\
        \normalsize\itshape
        $^2$ Department of Physics, Politechnika Gdanska, \\
        \normalsize\itshape
        Narutowicza 11/12, 90-952 Gdansk, Poland \\
        \normalsize
        E-Mail: \url{mcachor@pg.gda.pl}
        }
\date{}
\maketitle              
\begin{abstract} 
\noindent 
The mathematical formalism of quantum mechanics has been successfully employed in the last years to model situations in which the use of classical structures gives rise to problematical situations, and where typically quantum effects, such as {\it contextuality} and {\it entanglement}, have been recognized. This {\it Quantum Interaction Approach} is briefly reviewed in this paper focusing, in particular, on the quantum models that have been elaborated to describe how concepts combine in cognitive science, and on the ensuing identification of a quantum structure in human thought. We point out that these results provide interesting insights toward the development of a unified theory for meaning and knowledge formalization and representation. Then, we analyze the technological aspects and implications of our approach, and a particular attention is devoted to the connections with symbolic artificial intelligence, quantum computation and robotics.
\end{abstract}
\vspace{-0.2 cm}

\section{Introduction}
The use of the mathematical formalism of quantum mechanics as a modeling instrument in disciplines different from physics is now a well established practice and has historically been motivated by different reasons. Firstly, this is due to the flexibility and richness of quantum structures (vector spaces, inner products, quantum probability, quantum logic connectives, etc.). Secondly, there are two aspects that are seemingly characteristic of quantum entities, i.e. {\it contextuality} and {\it entanglement}, and that appear instead independently of the microscopic nature of these entities. Thirdly, the fact that since the fifties and sixties several effects have been recognized in a variety of areas, such as, economics, biology, psychology \ldots in which the application of classical structures (set theory, classical logic, Kolmogorovian probability, etc.) is problematical and generates paradoxes. The Allais \cite{allais1953} and Ellsberg \cite{ellsberg1961} paradoxes in economics, the conjunction fallacy  \cite{tverskykahneman1983} and disjunction effect \cite{tverskyshafir1992} in decision theory, the representation of concepts and the formalization of meaning in cognitive science \cite{oshersonsmith81}, are the most important examples of situations in which classical structures do not provide satisfactory results, but more general structures are needed. In particular, the impossibility of formalizing and structuring human and artificial knowledge slackened, notwithstanding the impressive technological success, in the development of some applied research fields, such as artificial intelligence and robotics.

The above difficulties led scholars to look for alternative approaches. Quantum mechanics then provided a fresch conceptual framework to address these problems in a totally new light. Hence, a {\it Quantum Interaction Approach} was born as an interdisciplinary perspective in which the formalism of quantum mechanics was used to model specific situations in domains different from the microscopic world. In particular, the new emerging field of Quantum Interaction focusing on the application of quantum structures to cognition has been named {\it Quantum Cognition} \cite{quantumcognition}. It is interesting, in our opinion, to dwell upon the main results obtained by the scholars involved in this Quantum Interaction Approach. We stress, however, that the following presentation does not pretend to be either historically complete or exhaustive of the various subjects and approaches that have been put forward, but it just aims to provide an overall conceptual background in which the approach on quantum cognition in which the authors of the present article and their collaborators are themselves intensively involved can be situated. 

The first insights came from psychology. In 1994 one of the authors and his collaborators proved that classical probability cannot be used to study a class of psychological situations of decision processes, but a more general probabilistic framework is needed \cite{aertsaerts1994}. In 2002 a contextual formalism generalizing quantum mechanics was worked out to model concept combinations \cite{gaboraaerts2002}. The SCoP formalism was successively improved and extended to provide a solution of the {\it Pet-Fish problem} \cite{aertsgabora2005a,aertsgabora2005b}. Since 2007 explicit quantum models in Hilbert and Fock spaces have been elaborated to describe experimental membership weights of concept disjunctions and conjunctions \cite{aerts2009a,aerts2009c,aertsaertsgabora2009,aertsdhooghe2009,aerts2007a,aerts2007b}. These models were then applied to study the Ellsberg paradox, the conjunction fallacy and disjunction effect in decision theory \cite{aerts2009c,aertsdhooghe2009,aertsdhooghehaven2010}, and a number of quantum effects, e.g., superoposition, interference, entanglement, contextual emergence, have been recognized in these effects. This fact led the authors to put forward the hypothesis that human thought presents two intertwined modes, one modeled by classical logic and the other mostly modeled by quantum mechanics \cite{aertsdhooghe2009}. 

Concerning cognitive models of knowledge representation, it was shown that modern approaches to semantic analysis, if reformulated as Hilbert space problems, reveal quantum structures similar to those employed in concept representation. In 2004 two of the authors recognized quantum structures in latent semantic analysis and in distributed representations of cognitive structures developed for the purposes of neural networks \cite{aertsczachor2004}. 

Interesting ideas came from information retrieval. In 2003 Dominic Widdows proved  that the use of quantum logic connective for negation, i.e. orthogonality, provided a more efficient algorithm than the corresponding Boolean `not' of classical logic for exploring and analyzing word and meaning \cite{widdows2003,widdows2006,widdowspeters2003}. In 2004 Keith van Rijsbergen claimed in his book that the Hilbert space formalism was more effective than an unstructures vector space to supply theoretical models in information retrieval \cite{vanrijsbergen2004}. Since then, several quantum effects have been recognized in information retrieval and natural language processing, e.g., superposition, uncertainty, entanglement \cite{licunningham2008}.

In 2006 Peter Bruza and his collaborators applied quantum structures to model semantic spaces and cognitive structures. More specifically, they undertook studies on the formalization of context effects in relation to concepts \cite{bruzacole2005}, and investigated the role of quantum structures in language, i.e. the entanglement of words in human semantic space resulting from violations of Bell's inequalities \cite{bruzaetal2008,bruzaetal2009}. 

In decision making important contributes were given by Jerome Busemeyer, Andrei Khrennikov and their collaborators. More specifically, in 2006 Busemeyer modeled the game theoretic variant of the disjunction effect on a quantum game theoretic model and used the Sch\"{o}dinger equation to describe the dynamics of the decision process \cite{busemeyerwangtownsend2006}. The proposed model is a part of a general operational approach of comparing classical stochastic models with quantum dynamic models, and deciding by comparison with experimental data which of both them has most predictive power \cite{pothosbusemeyer2009}. In 2008 Khrennikov presented a quantum model for decision making: he found an algorithm to represent probabilistic data by means of complex probability amplitudes, and used the algorithm to model the Prisoners Dilemma and the disjunction effect \cite{khrennikov2009}.

Quantum structures were hypothesized in the studies of holographic models of memory, which is an old research field in the psychology of memory. The metaphor which originally started this field is {\it holography}, that is, the observed fact that the human brain seems to have not really `places' for different functions ({\it holonomic brain theory}) \cite{pribram1971}. But, holography is a `wave effect related to electromagnetism', as is well known from physics \cite{gabor1968}. For this reason, some authors suggested that the results obtained in these holographic models of memory are due to an underlying quantum structure \cite{kanerva1998,plate2003,aertsczachordemoor2009}.

We will discuss the presence of quantum structures and their role in cognition, knowledge representation and information retrieval in a forthcoming paper \cite{aertsgaborasozzoveloz2011}. In the present paper we instead focus on the application of quantum structures to semantic analysis, artificial intelligence and robotics. More specifically, we resume in Sec. \ref{brussels} the main results that have been obtained by one of us in quantum cognition, including the hypothesis about the existence of a quantum layer in human mind. In Sec. \ref{ai} we instead explore how quantum structures can be successfully used to construct models in semantic analysis and symbolic artificial intelligence. Finally, in Sec. \ref{robotics} we investigate the links between our quantum cognition approach and quantum robotics, which is an emerging field that connects robot technology with quantum computation. We suggest, in particular, that macroscopic devices can be constructed which efficiently simulate quantum computers, thus avoiding the difficulties arising from the utilization of microscopic entities in quantum computation and robotics.

To conclude this section, we remind that the Quantum Interaction Group organizes each year an international conference on the Quantum Interaction Approach (see, \url{http://ir.dcs.gla.ac.uk/qi2007/, http://ir.dcs.gla.ac.uk/qi2008/, http://www-ags.dfki.uni-sb.de/~klusch/qi2009/} and \url{http://www.rgu.ac.uk/qi2011}) in which physicists, mathematicians, philosophers, psychologists, computer scientists meet to present and discuss the new results obtained in applying quantum structures to social, cognitive, semantic processes.

\section{Quantum structure in human thought\label{brussels}}
The proposal of using non-classical logical and probabilistic structures outside physics came primarily from an accurate analysis of the nature of the quantum mechanical probability model and of the difference between classical and quantum probabilities \cite{aerts1986,aerts1994,aertsaertscoeckedhooghedurtvalckenborgh1997,aerts1998b,aerts1999b,aerts2002}. This critical comparison led us to conclude that classical probabilistic structures formalize the subjective ignorance about what actually happens, hence they model only situations that admit an underlying deterministic process. However, it is well known that situations exist in quantum mechanics which are fundamentally indeterministic and cannot be explained in terms of lack of knowledge. Whenever this reasoning is applied to decision processes, one can see that human decision models are quantum in essence, because opinions are not always determined. This result has been shown by one of us by working out a quantum model for the decision process in an opinion poll \cite{aertsaerts1994}. But, the domain where classical set-theoretical based structures most maniflestly failed was concept theory and, specifically, the study of `how concepts combine'. This failure was explicitly revealed by Hampton's experiments \cite{hampton88a,hampton88b} which measured the deviation from classical set-theoretic membership weights of exemplars with respect to pairs of concepts and their conjunction or disjunction. Hampton's investigation was motivated by the so-called {\it Guppy effect} in concept conjunction found by Osherson and Smith \cite{oshersonsmith81}. These authors considered the concepts {\it Pet} and {\it Fish} and their conjunction {\it Pet-Fish}, and observed that, while an exemplar such as {\it Guppy} was a very typical example of {\it Pet-Fish}, it was neither a very typical example of {\it Pet} nor of {\it Fish}. Therefore, the typicality of a specific exemplar with respect to the conjunction of concepts shows a classically unexpected behavior. Since the work of Osherson and Smith, the problem has been referred to as the {\it Pet-Fish problem} and the effect has been called the {\it Guppy effect}. It can be shown that fuzzy set based theories \cite{zadeh01,zadeh02,oshersonsmith02} cannot model this `typicality effect'. Hampton identified a Guppy-like effect for the membership weights of exemplars with respect to both the conjunction \cite{hampton88a} and the disjunction \cite{hampton88b} of pairs of concepts. Several experiments have been performed (see, e.g., \cite{hampton01}) and many approaches have been propounded to provide a satisfactory mathematical model of concept combinations, but none of them provides a satisfactory description or explanation of such effects. Trying to cope with these difficulties one of the authors has proposed, together with some co-workers, a {\it SCoP formalism} which is a generalization of the quantum formalism \cite{gaboraaerts2002,aertsgabora2005a,aertsgabora2005b,aertsczachordhooghe2006}. In the SCoP formalism each concept is associated with well defined sets of states, contexts and properties. Concepts change continuously under the influence of a context and this change is described by a change of the state of the concept. For each exemplar of a concept, the typicality varies with respect to the context that influences it, which implies the presence of both a {\it contextual typicality} and an {\it applicability effect}. The {\it Pet-Fish problem} is solved in the SCoP formalism because in different combinations the concepts are in different states. In particular, in the combination {\it Pet-Fish} the concept {\it Pet} is in a state under the context {\it The Pet is a Fish}. The state of {\it Pet} under the context {\it The Pet is a Fish} has different typicalities, which explains the guppy effect. Inspired by the SCoP formalism, a mathematical model using the formalism of quantum mechanics, both the quantum probability and Hilbert space structures, has been worked out which allows one to reproduce the experimental results obtained by Hampton on conjunctions and disjunctions of concepts. This formulation identifies the presence of typically quantum effects in the mechanism of combination of concepts, e.g., contextual influence, superposition, interference and entanglement \cite{aerts2009a,aerts2009c,aertsaertsgabora2009,aertsdhooghe2009,aerts2007a,aerts2007b,aerts2010}. Quantum models have also been elaborated to describe the disjunction effect and the Ellsberg paradox, which accord with the experimental data collected in the literature \cite{aerts2009c,aertsdhooghe2009,aertsdhooghehaven2010}.

The analysis above allowed the authors to put forward the hypothesis that two structured and superposed layers can be identified in human thought: a {\it classical logical layer}, that can be modeled by using a classical Kolmogorovian probablity framework, and a {\it quantum conceptual layer}, that does not admit a Kolmogorovian probabilistic model. The latter mode can instead be modeled by using the probabilistic formalism of quantum mechanics. We stress, to conclude this section, that the thought process in the quantum conceptual layer is strongly influenced by the overall conceptual landscape, hence context effects become fundamental.

\section{Quantum structure in artificial intelligence\label{ai}}
Cognitive models of knowledge representation are relevant also from a technological point of view, for the representation of objects, categories, relations between objects, etc., play a central role in the development of artificial intelligence. In the last years techniques coming from quantum information theory have been implemented in the studies on semantic analysis and neural networks. In 2004 two of the authors proved that modern approaches to quantitative linguistics and semantic analysis, when reformulated as Hilbert space problems, reveal formal structures that are similar to those known in quantum mechanics and quantum information theory, hence in the quantum models on concept representation \cite{aertsczachor2004}. Similar situations are recurring in distributed representation of cognitive structures developed for the purpose of neural networks. Let us discuss two interesting aspects of these quantum approaches. 

Modern approaches to semantic analysis typically model words and their meanings by vectors from finite-dimensional vector spaces (see, e.g., latent semantic analysis \cite{landauer1998}). Semantic analysis is mainly based on text co-occurence matrices and data-analysis technique employing singular value decomposition. Various models of semantic analysis provide powerful methods of determining similarity of meaning of words and passages by analysis of large text corpora. The procedures are fully automatic and allow to analyze texts by computers without an involvment of any human understanding. The interesting thing is that there are strong similarities between latent semantic analysis and formal structures of quantum information theory. Latent semantic analysis is essentially a Hilbert space formalism. One represents words by vectors spanning a finite-dimensional space and text passages are represented by linear combinations of such words, with appropriate weights related to frequency of occurence of the words in the text. Similarity of meaning is represented by scalar products between certain word-vectors (beloging to the so-called semantic space). In quantum information theory words, also treated as vectors, are being processed by quantum algorithms or encoded/decoded by means of quantum cryptographic protocols. Although one starts to think of quantum programming languages, the semantic issues of quantum texts are difficult to formulate. Latent semantic analysis is in this context a natural candidate as a starting point for ``quantum linguistics''. Still, latent semantic analysis has certain conceptual problems of its own. As stressed by many authors, the greatest difficulty of this theory is that it treats a text passage as a ``bag of words'', a set where order is irrelevant. The difficulty is a serious one since it is intuitively clear that syntax is important for evaluation of text meaning. The sentences ``Mary hit John'' and ``John hit Mary'' cannot be distinguished by latent semantic analysis; ``Mary did hit John'' and ``John did not hit Mary'' have practically identical representations in this theory because `not' is in latent semantic analysis a very short vector. What latent semantic analysis can capture is that the sentences are about violence. We think that experience from quantum information theory may prove useful here. A basic object in quantum information theory is not a word but a letter. Typically one works with the binary alphabet consisting
of 0 and 1 and qubits. Ordering of qubits is obtained by means of the tensor product: we maintain that ordering of words can be obtained in the same way.

In 1990 Smolensky \cite{smolensky1990} proposed the introduction of tensor products of vectors to solve the so-called {\it binding problem}, i.e. how keeping track of which features belong to which objects in a formal connectionist model of coding. In the linguistic framework of semantic analysis the binding problem is equivalent to the problem of representing syntax. More specifically, one represents an {\it activity state} of a network by a vector (in a fixed basis), then a predicate $p(a,b)$, such as $eat(John,fish)$, is represented by the vector $r_1 \otimes a+r_2 \otimes b$, where the vectors $r_k$ represent {\it roles} and $a,b$ are {\it fillers}. A predicate is, accordingly, represented by an {\it entangled activity state}. It is important to note that tensor products are more `economic' than Cartesian products, because of the identifications $(\alpha|\psi\rangle)\otimes |\phi\rangle=|\psi\rangle \otimes (\alpha|\phi\rangle)=\alpha(|\psi\rangle \otimes |\phi\rangle)$, thus Hilbert (or Fock) spaces automatically perform a kind of dimensional reduction, which is the main idea of both latent semantic analysis and distributed representation. Furthermore, if one is interested in binding, more than ordering, words, then further compression is possible by employing bosonic or fermionic Fock spaces.

The above proposals on the advantages of using the quantum formalism in theories as semantic analysis and symbolic artificial intelligence find a straightforward theoretical support in our approach in cognitive science followed in \cite{aertsgabora2005a,aertsgabora2005b} and resumed in Sec. \ref{brussels}. Indeed, if the conceptual mode of human thought has a formal quantum structure, then it is natural to assume that the quantum formalism should be more successfully employed in cognitive disciplines. The same conclusion can be drawn if one assumes that the brain is a quantum device, as done in \cite{penrose1990}. But we stress that such an assumption is not needed in our approach (a similar remark will be made in Sec. \ref{robotics} with respect to quantum robotics).

\section{Quantum structure in robotics\label{robotics}}
The idea of a quantum robot meant as a complex quantum system interacting with an external environment through quantum computers was introduced by Paul Benioff in 1998 \cite{benioff1998}. Benioff undertook the study of quantum robots from a physical perspective. The first applications of Benioff's proposal to robot technology  are due to Daoyi Dong et al. \cite{dongchenzhangchen2006}. The model of a quantum robot suggested by these authors is made up of multi-quantum computing units, a quantum controller/actuator and information acquisition units. A quantum robot has also several learning control algorithms, including quantum searching aglorithms and quantum reinforcement learning algorithms. The standard problems afflicting classical robotics, i.e. robots' intelligence, sensor performance, speed of learning and decision making, are solved by using quantum sensors, parallel computing, fast searching and efficient learning from quantum algorithms. In particular, the authors point out the advantages in using Grover's search algorithm, which reduces the complexity of the search algorithm with respect to classical robots. 

We observe that the above insights and ideas rest on the possibility of constructing real quantum computers, implementing quantum operations on microscopic entities, and thus exploiting the computational advantages that quantum computation should guarantee over classical computation. It is however well known that several technical difficulties, besides conceptual hindrances, occur whenever one accepts to consider seriously the possibility of constructing a concrete quantum computer. The control and manipulation of individual quantum systems, the necessity of robustly representing quantum information, the actual feasibility in performing quantum algorithms, are examples of such difficulties. Hence, also the realizability of an efficient quantum robot strongly depends on these technological obstacles. 

Let us now come to our quantum cognition approach resumed in Sec. \ref{brussels}. Here, the fact that the formalism of quantum mechanics can be successfully employed to model concept representation, decision making and cognitive processes suggests that, conversely, the processes working in human mind have structurally a quantum nature. And this fact does not necessarily entail the compelling requirement that microscopic quantum processes occur in human mind. Indeed, following the quantum cognition approach, the hypothesis is rather that macroscopic processes can entail quantum structure without the necessity of the presence of microscopic quantum processes giving rise to these macroscopic processes. As a consequence of this hypothesis, one could maintain that human mind itself works as a system which is closer to a quantum computer than it is to a classical computer. It does not necessarily has to be equivalent with a quantum computer -- we believe it is not --, but entailing quantum structure gives it similar advantages in computing power than the one that quantum computers have over classical computers. This insight could explain, in particular, why artificial intelligence and robotics are still facing some fundamental problems, notwithstanding their impressive technological success: this is due to the fact that they use the paradigm of classical computation which is not powerful enough to perform the operations that the human mind is able to do. Let us finally remind that some of us have worked out macroscopic models ({\it connected vessels of water}, {\it quantum machine}, \ldots) which show a quantum behavior and exhibit typical features of quantum mechanical entities, i.e. contextuality, entanglement, violation of Bell's inequalitities, etc. \cite{aertsaertscoeckedhooghedurtvalckenborgh1997,aerts1998b,aerts1999b,aertsczachordhooghe2006}. This result is relevant in the perspective of quantum robotics because it opens up the possibility that the resources of quantum computation can be sought in other types of realizations than microscopic quantum entities and qubits. One could indeed envisage the possibility of elaborating (eventually complex) macroscopic devices which perform quantum algorithms, thus simulating quantum computers and exploiting the enormous extra power coming from quantum computation. In this way, the foregoing problems connected with the control of microscopic entities would be avoided and, better, the possibility of performing quantum computation by using only classical physics could potentially allow one to increase the resources of quantum computation itself.     

\section*{Acknowledgment}
This research was supported by Grants G.0405.08 and G.0234.08 of the Flemish Fund for Scientific Research.

\end{document}